\documentclass[11pt]{article}

\usepackage{tikz}
\usepackage{forest}
\usepackage{xcolor}
\usepackage{amsmath}
\usepackage{graphicx}
\usepackage{hyperref}
\usepackage{float}
\usepackage{placeins}

\usepackage{xcolor}
\definecolor{eggplant}{HTML}{6AA84F}
\definecolor{mygreen}{HTML}{F08700}
\definecolor{myred}{HTML}{F08700}

\usetikzlibrary{shadows,shapes,arrows.meta,calc,positioning,fit,backgrounds}
\usepackage[final]{acl}

\usepackage{times}
\usepackage{latexsym}

\usepackage[T1]{fontenc}

\usepackage[utf8]{inputenc}

\usepackage{microtype}

\usepackage{inconsolata}

\usepackage{graphicx}

\title{Test-time Scaling of LLMs: A Survey from A Subproblem Structure Perspective}

\author{
  \textbf{Zhuoyi Yang\textsuperscript{1}},
  \textbf{Xu Guo\textsuperscript{2}},
  \textbf{Tong Zhang\textsuperscript{2}},
  \textbf{Huijuan Xu\textsuperscript{1}},
  \textbf{Boyang Li\textsuperscript{2}}
\\
\\
  \textsuperscript{1}Department of Computer Science and Engineering, The Pennsylvania State University, United States \\
  \textsuperscript{2}School of Computer Science and Engineering, Nanyang Technological University, Singapore
\\
  \small{
    \textbf{Correspondence:} \href{mailto:boyang.li@ntu.edu.sg}{boyang.li@ntu.edu.sg}
  }
}

\begin{document}
\maketitle
\begin{abstract}
   With this paper, we survey techniques for improving the predictive accuracy of pretrained LLMs by allocating additional compute at inference time. In categorizing test-time scaling methods, we place special emphasis on how a problem is decomposed into subproblems and on the topological organization of these subproblems—whether sequential, parallel, or tree-structured. This perspective allows us to unify diverse approaches such as Chain-of-Thought, Branch–Solve–Merge, and Tree-of-Thought under a common lens. We further synthesize existing analyses of these techniques, highlighting their respective strengths and weaknesses, and conclude by outlining promising directions for future research.
\end{abstract}

\section{Introduction}
Test-time scaling (TTS) refers to the strategy of trading more computational resources for more predictive accuracy at inference time~\cite{brown2024large,openai2024reasoning,wu2024inference,deepseek-r1:2025}. By trading additional inference-time compute for accuracy, test-time scaling improves the performance of Large Language Models (LLMs) and Vision-Language Models (VLMs) while keeping model parameters unchanged, with demonstrated gains on complex tasks like ARC-AGI \cite{chollet2019:ARC}.

The most classic example of test-time scaling is arguably the Chain-of-Thought (CoT) technique \cite{wei2022cot}, which \emph{sequentially} generates a number of intermediate textual tokens describing the thinking process before predicting the final answer. These intermediate tokens can be shown to expand the range of problems solvable by a Transformer network with a constant number of layers~\cite{li2024chainthoughtempowerstransformers}.  

Besides the sequential execution of CoT, it is possible to organize the subtasks or subproblems differently. The central argument of this survey paper is that the subproblem structure is critical to performance, which is a principle well-known in theoretical computer science \cite{cormen01introduction} and exemplified by techniques such as divide-and-conquer or dynamic programming. Techniques, like Branch-Solve-Merge \cite{saha-etal-2024-bsm} and A-Decoding*~\cite{chatziveroglou2025adecoding} solve subproblems either in parallel or in a tree-search manner. 

In this paper, we survey test-time scaling techniques from the perspective of subproblem structures. A key insight we provide is that the identification and organization of the subproblems have important implications shared across problem domains (unimodal vs. multimodal) or LLM families (direct generation vs. retrieval-augmented\footnote{Retrieval-augmented Generation (RAG) \cite{NEURIPS2020_6b493230} may be seen as enforcing a task decomposition where the first subproblem is always retrieving from a database or a text corpus.}). To illustrate this point, we survey techniques across problem domains and LLM families and discuss their common strengths and weaknesses with the hope to inform future research. 

It is necessary to define the scope of this survey. We limit ourselves to methods that do not alter the parameters of the main LLM, but we do not rule out methods that finetune a small, auxiliary LLM, such as a learned heuristic function in tree search (Section \ref{sec:tree_structured_inference}). We exclusively focus on methods trade compute for accuracy, not methods that accelerate inference. Importantly, this survey primarily focuses on the subtask/subproblem structures rather than area- or domain-specific techniques, such as retrieval techniques in retrieval-augmented generation (RAG).

Further, it is beneficial to articulate the differences between TTS and a few related areas. TTS could be applied to improve LLM/VLM reasoning or agentic LLMs, but these two areas contain other techniques distinct from TTS. For example, reasoning could be improved by training-time techniques like reinforcement learning \cite{deepseek-r1:2025} and agentic LLMs contain topics such as societal simulation~\cite{park2023generative}. 

Our main contributions are as follows. First, we provide a detailed survey of TTS techniques from the perspective of subtasks structure, highlighting subtask decomposition as a paramount consideration in the design of TTS techniques. Second, we provide detailed discussion of the relative strengths and weaknesses of each decomposition strategy, which may guide proper TTS design for researchers and practitioners alike.

\begin{figure*}[t]
    \centering
    \includegraphics[width=0.8\textwidth, trim=0 30 0 0, clip]{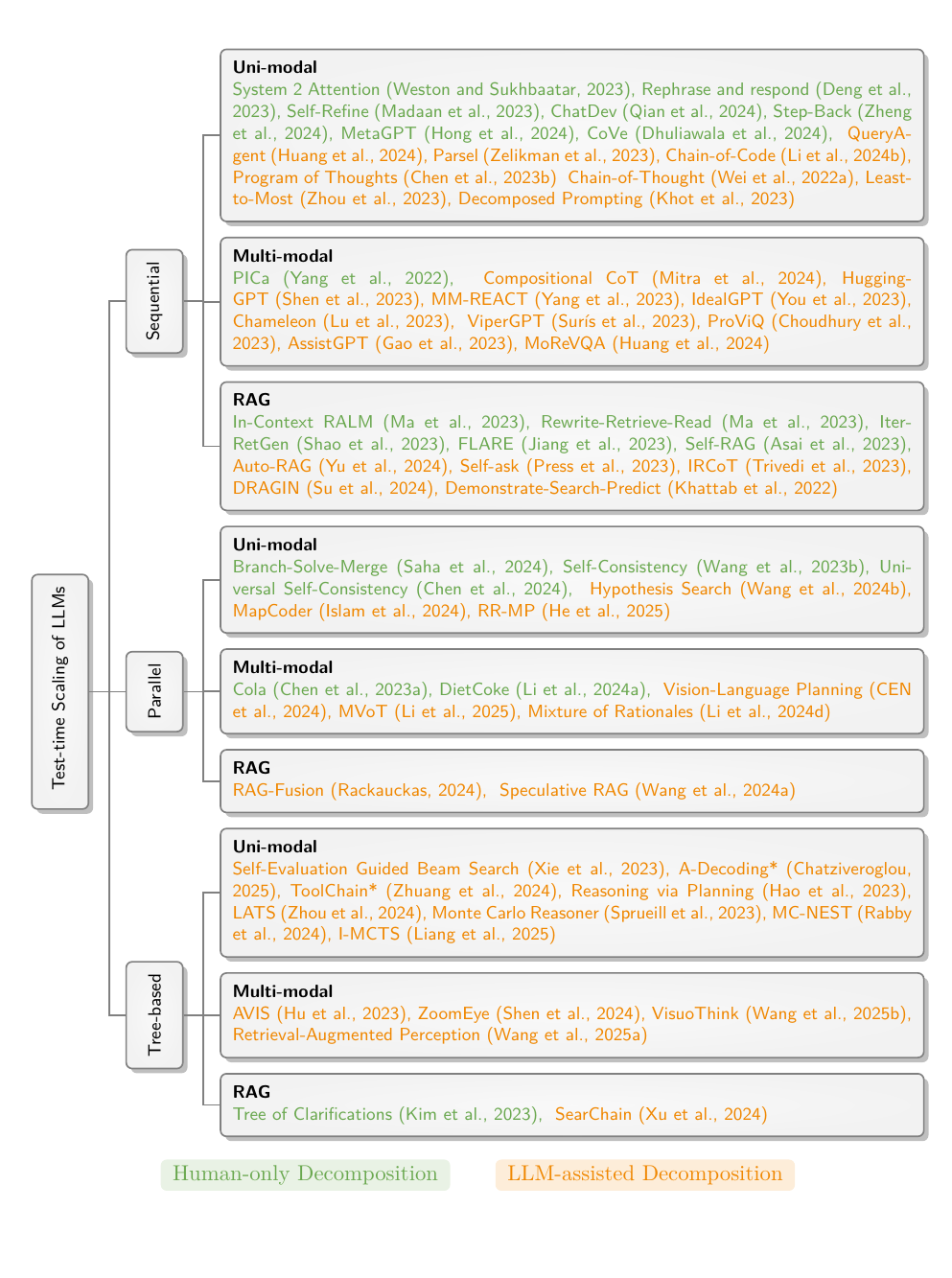}
    \caption{Test-time scaling methods categorized by reasoning path and decomposition method.}
    \label{fig:testtime-scaling}
\end{figure*}


\section{Task Decomposition}
\label{sec:task-decomposition}
The primary focus of the current survey is the decomposition of the target task into smaller, more manageable subtasks. Depending on the degree of automation, we categorize task decomposition strategies into two groups: human-only and LLM-assisted. 


\subsection{Human-only decomposition} 
In this strategy, the human designers provide a sufficiently detailed list or hierarchy of subtasks that do not need further decomposition. 
It is well suited to tasks with known processes and clear control flows, which may be crystallized from years of experience. One example is the software engineering process, for which~\cite{qian-etal-2024-chatdev} adopt the classic waterfall process of design, coding, code completion, code review, and testing. 

The human-only approach has a few advantages. 
First, a deterministic structure eliminates the need for the model to plan its steps, thereby improving inference efficiency and reducing variance. Second, it allows explicit designs that check and correct for known LLM weaknesses, to which some LLMs may be oblivious~\cite{gandhi2025cognitivebehaviorsenableselfimproving}. For example, we may enforce a self-verification step that rechecks an answer from an LLM or screens for unsafe outputs~\cite{xie2023selfevaluation}, or a step that removes irrelevant information that may mislead the LLM~\cite{deng2023rephrase}.
Similarly, Self-Refine~\cite{madaan2023selfrefine} introduces a subtask where the LLM criticizes and revises its own output. In the RAG paradigm, FLARE~\cite{jiang2023flare} detects low-confidence tokens in its predictions and revises them, whereas Self-RAG~\cite{asai2023selfrag} proposes evaluation subtasks that check the utility of retrieved documents and model predictions to the target problem. 

However, the rigidity of human-only decomposition limits the ability to customize the subtask hierarchy to specific inputs. This may be addressed by utilizing LLMs in task decomposition. 

\subsection{LLM-assisted decomposition} 

LLM-assisted decomposition allows the LLM to decompose some or all subtasks at inference time. This paradigm provides more flexibility and is suitable for tasks that require diverse, question-specific decomposition or have no obvious one-size-fits-all structures. A representative approach is Least-to-Most Prompting~\cite{zhou2023leasttomost}, which decomposes a complex problem explicitly into simpler subproblems in a question-specific manner. 
In contrast, Chain-of-Thought Prompting~\cite{wei2022cot} exemplifies an implicit decomposition approach, where the model directly generates step-by-step solutions in natural language without explicitly identifying intermediate subproblems. However, these reasoning steps still reflect an underlying problem structure.

Though LLM decomposition enables highly flexible and expressive reasoning, it may also produce unstable or suboptimal decompositions. For instance, the LLM may skip necessary subtasks or introduce irrelevant steps which may lead to incorrect solutions. A well-known fact is that chain-of-thought under reinforcement learning often produces unnecessarily long thoughts~\cite{wu2025less,yeo2025demystifying,kimi2025kimik15}. 

A hybrid decomposition strategy is also popular. In this strategy, human designers provide a clear outline for task decomposition, and the LLM subsequently refines portions of the high-level directions into more detailed subtasks at inference time, harnessing strengths of both approaches. 
Many variations of RAG follow this strategy; at a high level, there are only two tasks, Retrieval followed by Generation. But each of these tasks can be further decomposed into multiple queries~\cite{Rackauckas_2024} or iterative retrieval steps~\cite{trivedi-etal-2023-interleaving} by the model autonomously.

\begin{figure*}[t]
    \centering
    \includegraphics[width=1\textwidth]{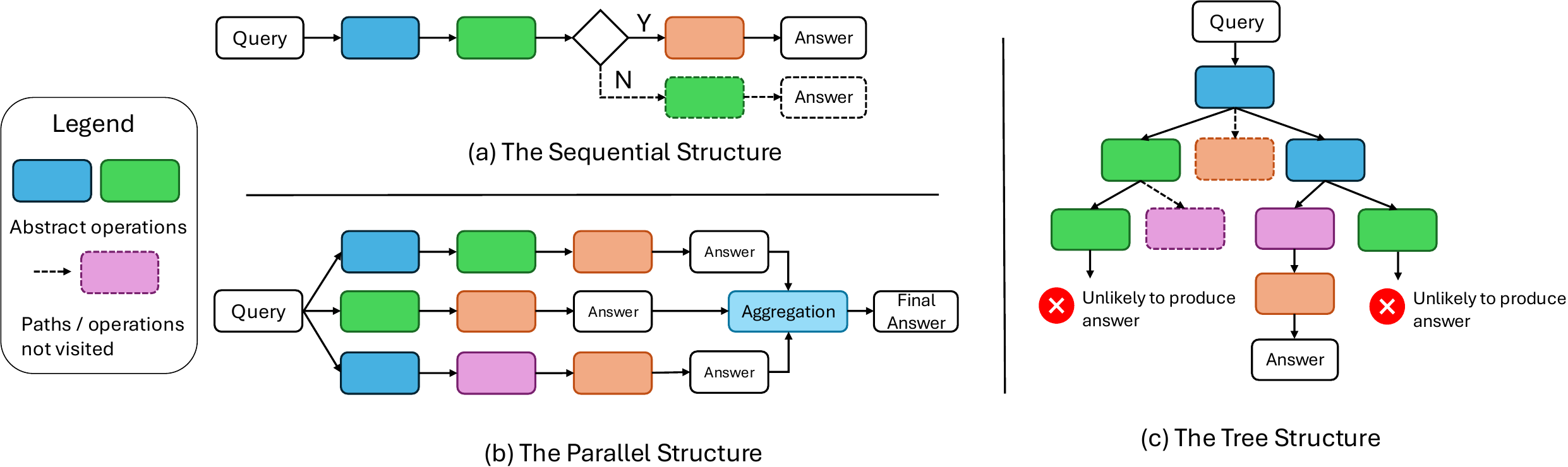} 
    \caption{An illustration of the three subtasks structures. In the sequential structure (a), there may be branching decision points, but we only visit one path leading to the answer without backtracking. In the parallel structure (b), we apply several paths simultaneously and aggregate the answers from all paths. In the tree structure (c), we may dynamically decide to jump from one path to another, but eventually find only one path to the answer. }
    \label{
    fig:dualcol1}
\end{figure*}

\section{Reasoning Paths}

In terms of the topological structure of subtasks, we identify four major types of subtask organization: sequential, parallel, and tree-structured. For each category, we start with a formal definition and then illustrate its applications in unimodal, multimodal, and retrieval-augmented generation (RAG) tasks. Each application scenario presents distinct challenges for subtask organization. Unimodal applications with textual input and output represent the most common use case. Multimodal applications integrate information across diverse modalities, introducing additional complexity. Even though RAG is not a specific modality setting, we place it in a separate subsection due to their internal similarity and prominent role in test-time scaling. 

\subsection{Sequential Structures}

\paragraph{Definition.} A sequential reasoning path organizes subtasks into a strict linear order, where later steps depend on the results of earlier ones. This structure ensures that intermediate outputs progressively guide the model toward a final solution. 

\paragraph{Unimodal Applications.}
As a pioneering work, \cite{wei2022chain} introduced Chain-of-Thought (CoT) prompting, which generates explicit intermediate reasoning steps in natural language, but not the explicit intermediate subtasks. This approach enhances LLM performance on numerical manipulations and logical inference tasks \cite{sprague2024cotcotchainofthoughthelps}. 
Under CoT, LLMs spontaneously determine the problem-solving process, including the granularity of problem decomposition and the methods for solving subproblems. 

The success of CoT has inspired subsequent research to explore more controllable ways of structuring intermediate steps to enhance accuracy and verifiability. For instance, Program of Thoughts \cite{chen2022program} and Chain of Code \cite{li2024CoC} translate natural language subproblems into executable programs, enabling more precise and verifiable subproblem-solving. Parsel~\cite{zelikman2023parsel} further enhances this idea by organizing problem-solving steps as structured trees of function-like elements, supporting fine-grained, interpretable computation. These methods offer a high degree of controllability by formalizing reasoning into syntactic structures.

Another direction focuses on improving the quality of LLM's output by embedding sequential feedback loops within LLMs' outputs.
Self-Refine \cite{madaan2023selfrefine} introduces a self-feedback cycle where the LLM critiques and iteratively improves its own outputs.
Step-Back Prompting \cite{zheng2024takestep} encourages the model to abstract from specific questions into high-level concepts, then reason based on these abstractions—reducing error propagation in complex inference.

\noindent\textbf{Multimodal Applications.} 
Vision-language tasks such as Visual Question Answering (VQA) are often handled in a straight from-model-to-answer approach. In contrast, PICa \cite{yang2022:PICa} leads the way for an test-time workflow. The process involves two sequential steps: (1) converting the image to verbal descriptions of the image, and (2) letting a text-only LLM answer the question based on the captions. Targeted captioning methods \cite{tiong-2022:PnP-VQA,Guo_2023_CVPR} further refine this idea by producing image captions specifically relevant to the question, yielding stronger performance than end-to-end trained VLMs such as Flamingo \cite{Alayrac-2022:Flamingo}. Scaling the number of captions at test time has been shown to improve answer accuracy \cite{wang-etal-2023-filling,xenos-etal-2023-simple}.

A second category integrates external tools or structured representations into the reasoning process. Compositional Chain-of-Thought \cite{Mitra2024CCoT} converts visual inputs into scene graphs (objects, attributes, and relations) and feeds them alongside the image as JSON into the VLM, facilitating structured intermediate reasoning. IdealGPT \cite{you2023idealgpt} follows a sub-question generation strategy akin to Least-to-Most prompting, using a VLM to sequentially answer each sub-question until it is confident in the final answer. Chameleon \cite{lu2023chameleon} further extends this pipeline by allowing the LLM to invoke external tools such as vision networks or Python functions to solve sub-questions. ViperGPT \cite{ViperGPT} and VisProg \cite{Gupta2022VisProg} compose visual reasoning pipelines by generating and executing modular Python programs using LLMs. These programs coordinate pretrained perception modules and logical operations, enabling interpretable and zero-shot solutions to complex vision tasks.

A similar sequential structure appears in multi-stage video reasoning. For instance, MoReVQA \cite{MoReVQA} first identifies key events in the video, grounds them in textual descriptions, and then applies memory-augmented reasoning before producing the final answer. Despite variations in implementation, all these methods share a sequential chain of intermediate results that guide the reasoning toward a final multimodal output.

\noindent \textbf{RAG Applications.}
Early methods in sequential RAG, such as In-Context RALM~\cite{ma-etal-2023-query}, adopts a single retrieval step during LLM inference to enhance factual accuracy. Rewrite-Retrieve-Read~\cite{ma-etal-2023-query} introduces a query rewriting step before retrieval to improve relevance. A more dynamic sequential RAG allows retrieval and generation to alternate in multiple iterations, with each iteration building upon the output of the previous one. Such methods usually incorporate an LLM-assisted reasoning process. Auto-RAG~\cite{yu2024autoragautonomousretrievalaugmentedgeneration} introduces a retrieval planner module that identifies unfulfilled information needs based on the retrieved information. The needed information is retrieved in the next iteration. Methods like IRCoT~\cite{trivedi-etal-2023-interleaving} incorporate Chain-of-Thought after retrieval to analyze and summarize external knowledge. 

In addition, some RAG methods such as Self-RAG~\cite{asai2023selfrag}, FLARE~\cite{jiang2023flare}, and DRAGIN~\cite{su-etal-2024-dragin} primarily rely on generation, only performing retrieval when external knowledge is needed or the model has low confidence in its output.

\noindent\textbf{Common Concerns and Mitigation.} 
The sequential nature of reasoning paths means that errors may propagate from earlier steps to later ones. To mitigate this, recent works propose strategies that inspect and refine intermediate steps to improve final output consistency. These strategies fall into three categories:

The first technique to reduce error propagation goes directly to the source: the user query, which can be ambiguous or poorly understood by the LLM. 
Rephrase and Respond (RaR) \cite{deng2023rephrase} addresses this by asking the model to restate or expand the user’s query in more explicit terms. This more transparent starting point supports more accurate subsequent reasoning steps. 
In multi-hop reasoning or STEM tasks, getting bogged down in details can confuse an LLM. Step-Back Prompting \cite{zheng2024takestep} encourages the model to abstract away specific details by deriving high-level questions and then revisit the original problem with these abstract questions to guide each reasoning step. This abstraction allows the model to see if the reasoning path is coherent at a conceptual level. 

The second technique attempts to let LLM to detect and correct its own errors. 
Self-Refine \cite{madaan2023selfrefine} prompts the LLM to generate an initial output and iteratively refine that output based on the model's own feedback. In each refinement round, the model pinpoints errors or ambiguities and then corrects them before moving on. To avoid producing plausible yet factually incorrect intermediate results, Chain-of-verification \cite{dhuliawala2024cove} generates additional “verification questions” that reflect on each critical fact of the chained thoughts. Each verification question checks a different piece of the initial draft independently, reducing the influence of prior biases or errors.  

The third technique seeks validation of intermediate results from external sources such as facts retrieved from a knowledge base. 
QueryAgent \cite{huang-etal-2024-queryagent} tackles hallucination by introducing rich environmental feedback, such as knowledge-base retrieval results and Python interpreter outputs, at intermediate stages. If an intermediate step is incorrect or incomplete, it applies self-correction. 
In multi-hop tasks, the model needs to piece together multiple facts. Self-Ask \cite{press2023selfask} narrows the “compositionality gap” by prompting the model to ask itself explicit sub-questions and answer them before formulating the final solution. The sub-questions can be validated separately with external tools (e.g., web search) for fact-checking. 

\subsection{Parallel Structures}
\noindent\textbf{Definition} 
Unlike uni-path reasoning, parallel multi-path reasoning explores several solution paths in parallel, where each path is instantiated at the beginning, proceeds independently without further branching, and is only aggregated at the end. This setting can be viewed as a form of ensemble learning, since ensemble methods rely on the diversity of candidates, generating varied intermediate answers across paths allows the model to compare or aggregate them, improving the chance of finding more accurate solution.

\noindent\textbf{Unimodal Applications} 
Among the most straightforward strategies is Self-Consistency \cite{wang2023selfconsistency}, which samples diverse chains of thought and selects the most frequent solution as the final output. Universal Self-Consistency \cite{chen2024universal} generalizes this idea to long-form answers by letting the LLM compare and pick the best answer from multiple candidates. In contrast, Branch-Solve-Merge \cite{saha-etal-2024-bsm} explicitly splits a complex task into multiple parallel subproblems and then merges the partial solutions into a coherent final output. Going a step further, RR-MP \cite{he2025rrmp} adds an explicit evaluation step after each path generation, where a reflection agent critiques and refines the preliminary outputs generated by a reactive agent. This dual-agent interaction ensures that each reasoning path undergoes iterative correction, mitigating hallucinations and degeneration-of-thought, while the multi-path design aggregates insights across diverse reasoning trajectories to produce more accurate and robust solutions.

Another interesting idea comes from Hypothesis Search \cite{wang2024hypothesis}, which generates multiple candidate programs and checks each program on exemplar input-output pairs. The system keeps performing LLM-powered self-improvement on each program until one program succeeds on all training problems or the maximum rounds of self-improvement is reached. In the code synthesis problem, MapCoder \cite{islam-etal-2024-mapcoder} refines multiple high-level plans simultaneously. 

\paragraph{Multimodal Applications.}

Ensemble learning benefits from multiple predictions that are decorrelated with each other~\cite{breiman2001random}. The multimodal setting naturally allows for decorrelated solution pathways specific to each modality. For example, Vision-Language Planning (VLP)~\cite{cen2024vlp}, designed for video question answering, features a visual pathway and a language pathway. The visual pathway samples frames from the video and the generated future video. The language pathway decomposes questions into sub-questions. Questions are answered based on outputs of both pathways. MVoT~\cite{li2024mvot} utilizes complementarity between modalities by mixing textual and vision information in CoT. It renders each textual thought as an image and feeding it back to the model to guide further reasoning. This technique simulates a human-like process of “imagining while reasoning,” and enhances performance in spatially grounded reasoning tasks. 

Some methods explicitly employ complementary solution strategies to enhance VQA performance. DietCoke~\cite{li-etal-2024-dietcoke} ensembles three distinct question-answering strategies: caption-based, short-form knowledge, and long-form knowledge, and selects the final answer using rationale-guided aggregation, demonstrating strong complementarity across decision contexts. Similarly, Mixture of Rationales (MoR)~\cite{li2024mor} generates short image descriptions with diverse foci, such as attributes, spatial relations, and contextual cues, decides if they are relevant, and fuses the relevant descriptions to answer the question. Cola~\cite{chen2023cola} uses an LLM to coordinate multiple vision-language models (VLMs), each with distinct reasoning strengths, enabling answer selection through inter-model complementarity. 

\noindent\textbf{RAG Applications.} RAG with a parallel reasoning path refers to methods where either retrieval, generation, or both are carried out on multiple paths in parallel. RAG-Fusion~\cite{Rackauckas_2024} decomposes the origin question into multiple retrieval queries, and fuses the retrieved documents in a single response generation step. In contrast, Speculative RAG~\cite{wang2024speculativeragenhancingretrieval} employs a single retrieval query but introduces multiple generation paths to generate diverse, parallel answer drafts. After that, an LLM verifies and selects the best draft for the final response. 

\noindent\textbf{Common Concerns and Mitigation.}
While parallel reasoning increases robustness by exploring and aggregating multiple solution paths, it introduces unique challenges. A key concern is incoherence or conflict between parallel paths---independently generated reasoning chains may reach inconsistent or incompatible conclusions, making it difficult to aggregate them into a reliable final answer.

One mitigation strategy is consensus-based filtering. Self-Consistency \cite{wang2023selfconsistency} relies on majority voting, mechanically choosing the most frequent answer, which demands exact matches among candidate responses and thus struggles with minor surface-form variations. Universal Self-Consistency \cite{chen2024universal} prompts an LLM to choose the most consistent answer from several answer candidates without explicit voting, overcoming strict reliance on exact matches in free-form generation tasks. However, both approaches assume correctness correlates with frequency, an assumption that may not hold if popular but spurious reasoning paths dominate.

VLP~\cite{cen2024vlp} proposes that the selection between direct aggregation and explicit voting should depend on the capability of the LLM/VLM. For highly capable LLMs/VLMs like GPT4-V, VLP allows them to directly synthesize information and generate a free-form answer. Conversely, for LLMs/VLMs with limited capabilities, VLP implements a multi-round conversational strategy that incorporates a voting mechanism. Experimental results indicate significant contribution of the aggregation technique to performance and the potential for further research. 

Another challenge is resource allocation. Parallel reasoning is compute-intensive, and blindly expanding all paths can be inefficient. Some systems, such as Speculative RAG \cite{wang2024speculativeragenhancingretrieval}, mitigate this by using lightweight models to pre-screen candidates, reserving large model inference for only the most promising paths. Together, these techniques aim to retain the diversity benefits of parallel reasoning while optimizing compute usage.

\subsection{Tree Structures}
\label{sec:tree_structured_inference} 

\paragraph{Definition} A tree-structured reasoning path arranges subtasks in a branching hierarchy. There are two typical tree structures. In the first, more popular, structure, similar to path finding~\cite{hart1968formal}, a solution can be formed from only one path from the root to a leaf node~\cite{yao2023tot,10.1145/3589334.3645363}. In this tree type, each node represents a (partial) solution to a subtask, and children of a parent represent competing solutions.  

In the second tree structure, similar to hierarchical task networks~\cite{Ilche2015HTN}, the final solution requires aggregation of information from all leaf nodes~\cite{kim-etal-2023-tree}. Here the leaf nodes represent subtasks that must be combined to produce the final result. As all nodes of the tree must be visited, the order of visiting does not make significant differences. In most cases of LLM reasoning, the parent nodes are progressively decomposed into child nodes as reasoning unfolds, and typically only one branch is further explored toward the final solution, while other branches—corresponding to incorrect or suboptimal reasoning paths—are pruned or discarded. Therefore, most tree-based methods belong to the first type of trees, while in this section’s RAG part, we will introduce a method \cite{kim-etal-2023-tree} that extends the idea of the second tree structure.

For the first type of trees, how to traverse the tree to find or assemble a solution carries great importance. Classical tree traversal algorithms include depth-first search (DFS), breadth-first search (BFS), and informed heuristic search. 
DFS and BFS suffice for small trees, whereas informed search algorithms such as A*~\cite{hart1968formal} and Monte Carlo Tree Search (MCTS)~\cite{coulom2006MCTS} are more suitable for large trees, as they prioritize nodes based on estimates of how good each node is. 
In A*, each node is ranked using a combination of historic cost and an estimated future cost to the goal---traditionally computed via hand-crafted functions, but increasingly learned by neural networks~\cite{silver2016mastering,gupta2024training}. 
MCTS estimates node values through randomized simulations.

Compared to sequential and parallel structures, tree structures offer greater flexibility. 
They can dynamically decide which subtask to solve next, backtrack to explore alternative paths, and prune unpromising nodes to improve efficiency.

\paragraph{Unimodal Applications.}

Self-Evaluation Guided Beam Search \cite{xie2023selfevaluation} adopts beam search as its traversal strategy---a method that combines features of both BFS and heuristic search. It selects a fixed number (denoted as $K$) of nodes with the highest heuristic score at each tree level. Only the selected nodes get expanded---their children are generated. Next, it selects $K$ highest-scoring nodes among the children and the process repeats. Crucially, the tree node evaluation involves a quality score from the LLM alongside the probability of the generated partial thought.

Tree of Thoughts (ToT) \cite{yao2023tot} is well-known for formulating CoT generation as a tree search problem and proposes two traversal techniques. The ToT-BFS algorithm is similar to beam search; the ToT-DFS algorithm keeps expanding the current node while discarding any children whose evaluation score falls below a predefined threshold.

The aforementioned approaches still assume a manageable tree size or a bounded beam width. For larger search spaces, an informed search, such as A* or MCTS, becomes necessary. A* contains two critical functions, $g(\cdot)$ and $h(\cdot)$, which are the historic cost and the estimated future cost (i.e., distance from current step to goal), respectively. In A-Decoding*~\cite{chatziveroglou2025adecoding}, $h(\cdot)$ is the evaluation of each partial CoT candidate by an LLM, where a higher value indicates a worse candidate. To preserve the optimality guarantee of A*, the historic cost $g(\cdot)$ accumulates all the increase in $h(\cdot)$ along the trajectory but not any decrease. ToolChain*~\cite{zhuang2024toolchain} applies A* to generate a sequence of API calls, and collects successful API sequences in history. $g(\cdot)$ contains two parts: the cost of the most similar plan in the collected history, and the frequency that the LLM generates the current action from the current plan in repeated simulation. $h(\cdot)$ also contains two parts: positions of the current action (or the most similar action) in all historical plans, and the length of an LLM-generated future plan from the current step.

Monte Carlo Tree Search (MCTS) employs four stages: (1) selection, which selects a tree node to expand; (2) expansion, which generates children for a tree node; (3) simulation, which evaluates a node by rolling multiple random action sequences; and (4) backpropagation, which updates statistics at existing tree nodes with information acquired in simulation. A prototypical method combining LLM and MCTS is Reasoning via
Planning (RAP) \cite{hao-etal-2023-rap}. To solve a planning problem, RAP utilizes an LLM to generate both the actions and the state reached after applying each action. The states become tree nodes. In expansion, it generates actions and state using the LLM. In simulation, the rewards are computed from LLM-predicted action likelihoods, LLM self-evaluation, and human-specified domain-specific heuristics. In selection and backpropagation, RAP follows classic MCTS.  LATS \cite{zhou2024lats} is largely similar to RAP, but introduces a self-reflection stage where the LLM reflects upon a failed terminal node and generates text to describe the reason for failure. The generated text becomes part of the LLM context in subsequent search steps in order to prevent similar mistakes. LATS also employs external environments such as a Python compiler for feedback. 

The simulation stage of MCTS is particularly expensive for LLMs as it may involve many LLM inference calls. As a result, some techniques, such as Monte Carlo Reasoner
~\cite{sprueill-etal-2023-monte} and MC-NEST~\cite{rabby2024mc-nest} directly ask an LLM to evaluate the current tree node instead of performing random rollouts. These methods may be considered as simplified MCTS without the simulation stage.  Additionally, some methods combine LLM evaluation and rollout. For example, when faced with multiple leaf nodes that are eligible for rollout, I-MCTS~\cite{liang2025imcts} selects one node for rollout using an LLM value model. 

\paragraph{Multimodal Applications.}
There are a diverse set of tree-based techniques for multimodal applications. AVIS~\cite{hu2023avis} searches for a sequence of API calls, such as retrieval for similar images, or selection of objects on the image, for visual question answering. Each API call adds more information to the QA context. The system employs an LLM planner that determines the next API call to try, and backtracks if the API results are deemed uninformative. The LLM answers the question when it determines enough information has been collected by API calls in ancestor tree nodes. 

In contrast to the purely textual Tree-of-Thoughts, multimodal applications can adopt multimodal reasoning states as tree nodes. For example, in VisuoThink~\cite{wang2025visuothink}, each tree node contains the result from previous reasoning actions. The result may include both visual and textual information. To solve a geometry problem, the algorithm follows a Thought–Action–Observation cycle: (1) the Thought phase generates the next action, such as adding a new line as visual aid; (2) the Action phase executes the action and obtains a new visual state; (3) the Observation phase deduces new facts from the new visual state and incorporates them into the resultant node. VisuoThink performs MCTS as the tree search method.

Another possibility is build tree nodes with purely visual information. In ZoomEye~\cite{shen2024zoomeye}, a VQA method, the root node is the full high-resolution image and each child node is a zoomed-in subregion of its parent. It employs an MLLM-guided heuristic search, ranking candidate regions based on model-estimated confidence and simulating human-like zoom-in/zoom-out actions to progressively locate visual evidence relevant to the task. Retrieval-Augmented Perception~\cite{wang2025rap} builds a tree to gradually remove image patches irrelevant to the question. The image is first divided into a fixed grid of patches. As we go deeper in the tree, more image patches are discarded. A tree node contains a partial image that preserves the spatial relations of selected patches. The tree search is performed using A*, where $g(\cdot)$ is the average question-relevance score over selected image patches, and $h(\cdot)$ is calculated as one minus the LLM-estimated probability that the question can be answered from the current tree node. 

\noindent\textbf{RAG Applications.} Tree-structured RAG are under-investigated in the literature, and we discuss two representative methods. For each question it receives, SearChain~\cite{10.1145/3589334.3645363}, generates a sequence of sub-questions, which are a sequence of tree nodes from the root to a leaf. Each sub-question is answered by both an LLM directly and from retrieved documents. If the two answers disagree on any sub-question, this indicates the sub-question may be ill-posed or not answerable, and the system backtracks and generates a new sequence of sub-questions starting from its parent. Tree of Clarifications~\cite{kim-etal-2023-tree} resolves ambiguity in an original question (AQ), which is the root node. From a parent node, the RAG process generates several possible clarified questions (usually with more details), forming the children nodes. If the clarified question becomes irrelevant to AQ, it can be pruned by LLM self-verification. To form a final answer, the system aggregates answers to all questions in the tree. 

\noindent\textbf{Common Concerns and Mitigation.}
We identify three unique challenges of tree-structured reasoning. First, to allow back-tracking and pruning, the tree structures stores many nodes in memory, incurring significant space cost. Mitigation typically involves pruning or restricting tree growth. For example, ToT discards nodes with low evaluation scores~\cite{yao2023tot}, and \cite{xie2023selfevaluation} only expand the top-$K$ nodes at each level. Additionally, informed search such as A* and MCTS considers the future potential of each node, on top of its current estimated quality, leading to more intelligent node selection. However, they also require reliable evaluation heuristics or costly rollouts. 

The second challenge is the reliability of the node evaluation, which is often performed by an LLM. However, LLMs may suffer from certain biases that produce inaccurate evaluations \cite{wataoka2025self, ye2024just}. Using the same LLM that generates the tree nodes as the evaluator, like in RAP~\cite{hao-etal-2023-rap}, may suffer from the self-preference bias, where the LLM prefers text that it believes to have low perplexity \cite{wataoka2025self}. Using a different LLM, like in A-Decoding*~\cite{chatziveroglou2025adecoding}, may alleviate this bias but not other types of biases. Mitigation includes combining self-evaluation with external signals such as execution feedback~\cite{zhou2024lats} or retrieved evidence~\cite{10.1145/3589334.3645363}; or ensembling multiple evaluations, as in to Universal Self-Consistency~\cite{chen2024universal}; or delaying aggressive pruning until later search stages when more evidence is available.

A third challenge is the high computational cost of repeated rollouts in MCTS to estimate node values. Some methods, such as I-MCTS~\cite{liang2025imcts}, replace full rollouts with a direct LLM evaluation. Others~\cite{li-ng-2025-think} approximate rollout outcomes using lightweight models or heuristic functions, and reserve expensive LLM inference for only the most promising candidates. These strategies preserve the exploratory benefits of MCTS while substantially reducing computational overhead.

\section{Future Directions and Conclusions}
With this paper, we review test-time scaling methods decompose a difficult task into subtasks and reasoning paths, and discuss the trade-offs in their design. We identify several promising directions that have not been fully explored at the time of writing. 

\noindent\textbf{Meta-Reasoning: Learning to Select Reasoning Strategies.} Subtask selection and discovery remain challenging, especially when inputs are diverse: it is unclear whether to decompose, to what granularity, and whether the sequential, the parallel, or the tree-based structure provides the best performance. Most existing work relies on zero-shot prompting or supervised fine-tuning (e.g., Least-to-Most \cite{zhou2023leasttomost}), but leveraging meta-learning for “learning to reason” could be fruitful.

\noindent\textbf{Efficient Multi-Path Reasoning.} Parallel or tree-based reasoning extends sequential methods in order to improve robustness and accuracy, but also increase inference cost. Parallel approaches such as Speculative-RAG \cite{wang2024speculativeragenhancingretrieval} reduce cost by using lightweight models on branches, while tree-based methods like I-MCTS \cite{liang2025imcts} mitigate rollouts through value estimation. However, efficiency remains an important open problem for future work. 

\noindent\textbf{Tree-Based Reasoning in Multimodal and RAG Systems.} Tree-based reasoning offers great flexibility in exploring diverse reasoning paths, yet its applications in multimodal understanding and retrieval-augmented generation (RAG) remain limited. Extending tree search to multimodal and RAG systems could unlock richer subtask structures and more reliable inference.

We hope this survey provides a unifying perspective that builds a foundation, from which more efficient, reliable, and generalizable reasoning systems can be designed.

\bibliography{ref}

\end{document}